# Can a Chatbot Determine My Diet?: Addressing Challenges of Chatbot Application for Meal Recommendation

Ahmed Fadhil
*University of Trento*
*Trento, Italy*
*fadhil@fbk.eu*

*Abstract*—Poor nutrition can lead to reduced immunity, increased susceptibility to disease, impaired physical and mental development, and reduced productivity. A conversational agent can support people as a virtual coach, however building such systems still have its associated challenges and limitations. This paper describes the background and motivation for chatbot systems in the context of healthy nutrition recommendation. We discuss current challenges associated with chatbot application, we tackled technical, theoretical, behavioural, and social aspects of the challenges. We then propose a pipeline to be used as guidelines by developers to implement theoretically and technically robust chatbot systems.

*Keywords*-Health, Conversational agent, Recommender systems, HCI, Behaviour Change, Artificial intelligence

## I. INTRODUCTION

With the current era of connectivity and technological innovation, we're able to find, share and experience with our device and other users like never before. With services like Siri and Google Now becoming widely used, users are no strangers to interacting with machines designed to act-human. Recently, a big step forward happened with the rise of conversational interfaces. As this technology grows and develops, its reasonable to expect chatbots to be more sophisticated and intelligent. These AI powered technologies will ultimately do an increasingly impressive job of simulating interaction with a living, breathing virtual assistants.

There are already chatbots serving in certain domain to accomplish complex tasks or support humans to achieve high efficiency at their activity. Several platforms, including Facebook, KiK, Telegram and Slack are already hosting chatbot applications with several purposes, ranging from e-commerce, health, entertainment, productivity and lifestyle promotion.

Chatbots can provide users with health information and diet plans. There are a range of bots available depending on the type of diet and exercise plans the user needs, such as Health Hero[1], Tasteful Bot[2] and Forksy[3] all build for Telegram and Slack platforms. All these bots have the goal of helping users change unhealthy eating habits and give insights into their own eating habits to help them transform. Chatbots are also effective with exercise plans, fitness tips and monitoring. It's easy to provide this kind of advices without any physical interaction. These bots focus on motivation and providing everyday information to users. They trigger the user into the activity via notifications or rather automatic messages. They're giving a little piece of inspiration and reminders to help users stay on track. Chatbots can even determine user diet and help them to improve it to some extent. One point to consider is that they don't substitute health professionals, and they almost all tell you to seek the advice of your doctor or a health professional. However, for a healthy person interested in going vegan, for example, then bots have the capacity to provide you with all the information that you need. Finally, chatbots can track user health and determine if they are successfully exercising. One major reason why people seek trainers is that they don't exercise correctly. A trainer can help align your body correctly, teach you about reaching high intensity workout levels and show you exactly how to do exercises in a way that will get you the results you're looking for. This task is difficult for bots to tackle since they're purely chat-based and can't physically adjust the person they're assisting. Hence, in some domains, such as exercise, a real person is still needed [1]. Chatbots are developed by humans, and humans aren't perfect. According to BotAnalytics[4], nearly 40% of users stop talking to a bot after the first message and another 25% are gone after the second one. With this article, we investigate the application of chatbot systems in meal recommendation and lifestyle promotion. Our focus is to highlight the behavioural, theoretical, technical, design and even logical-flow challenges associated with building a robust conversational interface systems. The scope of the paper will tackle the issue of meal recommendations and the challenges associated with building a chatbot system to provide the necessary support within such system, since it's definitely an area of growing prevalence in the bot world. For example, chatbots that are able to give details on diet options, diet plans, restaurant recommendations, fitness tips, exercise regimes, ongoing motivation. Finally, we will provide some insights on the current state of advancement in conversational interfaces and discuss with a use-case scenario important factors to consider when designing, implementing and testing such systems. Most importantly, we believe such systems will create a new paradigm for human-bot interaction. We believe that such an approach is feasible and will lead to more effective user engagement with health and lifestyle promoting applications.

---

[1]https://fa.ebotstore.com/bot/a02sjhzq9-health-hero-slack
[2]http://www.marketwired.com/press-release/healthy-eating-app-tasteful-gets-personal-with-new-chatbot-feature-2127056.htm
[3]https://getforksy.com/
[4]https://botanalytics.co/

II. AI-BASED LIFESTYLE PROMOTION APPROACHES

Health and fitness are two extremely dominant terms in the modern world and one thing is abundantly clear. There are endless lists of diet types and exercise plans and weight loss tips provided by several applications. And this was perhaps partly because of the increasing amounts of people struggle with diet and weight based issues and obesity is more deadly than ever in the whole world [2]. This is an indication of the endless amount of people looking for solutions, ideas and plans to improve their health and fitness or simply seek out new diets, change up the way they eat or purse a healthier lifestyle. To determine a bot can function as a personal trainer or health and fitness advisor, we should test it agains some criteria. These criteria range from the way the bot triggers behaviour change, such as engagement, considering user need and burden. In designing for behaviour change, Grolleman et al., [3] described the requirements and design for building an embodied conversational agent as a virtual coach to support people with smoking cessation. The design focused on the most prominent nonverbal channels. The study suggested more of work on the improvement of natural language generation algorithms to enrich the functionality and humanness of the virtual coach. Research in persuasive technologies and the associated application to change an attitude or behaviour in a predetermined way is showing the potential to assist in improving healthy living, reduce the costs on the health care system, and allow the aged to maintain a more independent life. Embedding persuasive techniques in chatbot behaviour can enhance the bot efficiency in behaviour change. A study by Chatterjee et al., [4] presented a framework that can guide researchers in comprehending more effectively the work being done in persuasive technology. The research goal is to provide greater understanding by addressing the challenges that lie ahead when designing and using persuasive technologies in healthcare.

The technicality behind the bot architecture could be simple or sophisticated. For example, the chatbot system could use machine learning, neural networks, wearables or IoT (Internet of Things) devices to handle tasks, or it could be rule-based and use finite state machine to accomplish rather simpler tasks. A work by Shawar [5] described a software to machine-learning conversational patterns from a transcribed dialogue corpus to generate chatbots speaking various languages. The work presented a program to learn from spoken transcripts of the Dialogue Diversity Corpus of English, the Minnesota French Corpus, the Corpus of Spoken Afrikaans, the Qur'an Arabic-English parallel corpus, and the British National Corpus of English. Two goals were achieved, the ability to generate different versions of the chatbot in different languages and the ability to learn a very large number of categories within a short time.

Although designing CUIs seems simple, however its bounded by handling several complex tasks running behind the system that provide the respond to various user requests. Designing CUI is completely different than the well-known GUI system. Moreover, migrating services from a web or mobile application into a conversational application could get extremely complex. There is no clear way on how to map various UI elements and achieve a UX that fits with the new context. Perhaps, it's also necessary to consider the elements to keep within the CUI or introduce new once whenever necessary. Ait et al., [6] discussed the text-based and voice based interactions in a chatbot, then highlighted the Automatic Speech Recognition or Speech-To-Text and Text-To-Speech layers on top of a text-based interaction layer. The paper aimed to provide a text-based interface to a structured dataset with a clear focus on getting domain-specific actions executed while maintaining a smooth communication with the users.

With the rise of smartphone, IoT and Messaging platforms, speech and language processing, chatbot systems are starting to appear feasible. However, adapting a one-size-fits-all approach can't achieve much when we deal with specific domains and their specific requirements. Designing patterns for specific domains is essential to increase the possibility that the chatbot can achieve the goal set. Therefore, focusing on the context, content, user-bot interaction and adapting pattern matching techniques to design sophisticated user-bot conversations can enhance the task accomplishment [7].

Even with advanced AI platforms, there is still room for improvement. Chatbot designers have to craft the logical steps to follow so to achieve the goal set for the bot without falling into dead-ends. The logic behind the chatbot system depends a lot on the domain, its expected users, its behaviour and the conversation flow between the user and the chatbot. Among the available techniques used in crafting the chatbot logic are simple approaches, such as decision trees, or more complex, such as custom natural language processing (NLP) and machine learning (ML) components. Bhatia et al., [8] presented a textual reply generation model by taking into account the content preferences of the user, the idiosyncrasies of their conversational style, and the structure of their social graph. The study developed two types of models for personalised user interactions: a content-based conversation model, which makes use of location together with user information, and a social-graph-based conversation model, which combines content-based conversation models with social graphs.

Finally, a less technical term in chatbot design is the language used for interaction. Chatbots should have dialogue capabilities to respond to complex tasks and user interactions. That said, having intent detection capabilities, dialogue system and strong NLP component can increasingly enhance the linguistic capabilities for the bot. There already exists platforms to provide specific linguistic capabilities for chatbots, either to strengthen their text or speech capacities, to increase their behavioural AI, to analyse various user sentiments based on their conversation, and hence provide a tailored respond for user request. Natural language processing employs computational techniques for the purpose of learning, understanding, and producing human language content. Hirschberg et al., [9] described the challenges in creating spoken dialogue systems and speech-to-speech translation engines, mining social media for information about health and identifying sentiment and emotion toward products and services. A similar work by Andrews et al., [10] discussed applying the persuasive argumentation theories to human-computer dialogue management to achieve comfortable experience for the user. The study analysed different aspects of persuasive communication needed for healthcare advising and how to implement them to produce efficient persuasion. Another study on health by Crutzen et al., [11] investigated an artificially intelligent chatbot that answers questions about sex, drugs, and alcohol is used and evaluated by adolescents, in comparison with information lines and search engines. The chatbot reached high school attendees in general and not only adolescents with previous experience related to sex, drugs, or alcohol; this is promising from an informed decision-making point of view. In the context of diet and meal recommendation, Liu et al., [12] presented "CityBrowser", a conversational dialogue system which allows users to inquire information about restaurants in China. The work focused on the language specific modifications to the original English system. The findings showed the possibility to scale the system infrastructure for multilingualism.

III. PERSISTING CHALLENGES

In this section, we highlight the various theoretical, technical, behavioural, design, pattern, logical and linguistic challenges associated with chatbots systems intended for food recommendation and lifestyle promotion. Below we provide seven major domains where challenges associated with chatbot systems persists.

- **Theoretical Foundations:** Existing bot systems are rarely based on theoretical foundations and are generic in their domain focus. There is a need to consider theoretical models for chatbot systems. These models are essential to understand the theory behind user motivation, engagement, the perceived usefulness of the chatbot, conversation matching and cultural acceptance of the conversation tune. Basing the chatbot on a theoretical model will help match against the form of the incoming speech acts and generate possible underlying intentions. An approach example is the Schutz von Thun's four layer model [13] where communication and conversation has a sender and a receiver exchanging messages that contain elements on four layers (Factual information, Relationship, Self-revelation and Appeal).
- **Behavioural Change Techniques:** Although this is related to the theoretical foundations, still behaviour change techniques are more focused on how to discover a certain behaviour and aim to change it via behaviour change techniques. To understand how to engage users with the chatbot system, we first need to define the user behaviour we intend to tackle, then which behaviour change techniques are associated with positive changes in this behaviour. For example, if the behaviour is improving diabetes condition, then the behaviour technique should consider physical activity, diet and body mass index (BMI). Among the research validated behaviour change techniques, we mention the Persuasive System Design PSD [14] which is applicable to systems designed to form, alter or reinforce attitude, behaviour, or a compliance act without deception, coercion or inducements. This model provides principles (Primary Task Support, Dialogue Support, System Credibility Support, and Social Support) to consider when designing a persuasive technology system. Designers have to understand that processes of attitude formation and change are not identical, nor are their outcomes.
- **Technical Limitations:** Conversational AI is an emerging domain and has several technical limitations that are essential when building an AI powered chatbot. Chatbots are of two types:

    *Rule-based Bots* - These are the most common bots in consumer facing businesses, in the short-term. These Bots are easiest to build with less infrastructure than their more complicated sister - AI Bots. They are built to provide linear and single dimensional support for example, customer service. Using a finite-state systems or a decision tree are typical fixed system initiative used in building rule-based bots. At each point in the dialogue, the user must answer the specific question the system asks. As the task becomes more complex, these strategies become less and less useful. They have a specific technology and architecture requirement and are quicker and easier to build, deploy and implement, but therefore cheaper, less scalable and less robust with lots of potential down side for the long term for example with reliability, analytics, learning, less of an investment, more of a cost and liability.

    *AI-based Bots* - These are similar to rule-based bots, except that they have different tech requirements de-

manded to support a chatbot through to more advanced levels, as well as ongoing support, is the artificial intelligence component and Natural Language Processing capability. Artificial Intelligence bots provide an overall function far greater than a mere chatbot and require a brain and a Natural Language Processor (NLP) to convert and harness the outcomes of conversations, apply some intelligence and convert that knowledge through learning into powerful decision making capability, over the long term. With the rise of platforms to deliver precise Platform as a service (PaaS) solutions specifically around chatbots, building an intelligent bot for a specific domain with certain behaviours and capabilities became increasingly easy and efficient. Among the averrable techniques to consider are machine learning components, neural networks in meal recommendations and REST APIs.

- **UCD and UX Design:** The familiar design patterns used in GUIs are obsolete and do not work in conversation-driven interfaces. As mentioned above, before developing a chatbot one should ask the purpose and benefit of the chatbot for the user. The UCD and UX Design put focus on the user and the use a product or service generates. A set of guidelines often are required when it comes to UX design for CUIs. UX advocates often highlight the need for usability as a core element of good design and a good UX. The issue with these approaches is that they mainly focus on usability. But the issue often lies in their rigid or isolated application, ignoring other aspects. Good usability might be good for users with a rational, specific and task-oriented goal. But this might totally neglect the overall experience of a product. As earlier stated, some products might exclusively aim at fulfilling our life with fun, enjoyment, entertainment. Chatbots have specific UX requirements related to the way the information is presented, the mean of interaction, whether its text, buttons or speech. Moreover, UX designers have to pay attention to the way the bot presents the content to the user, for example, is the text too long or is the image too small and find. And hence enhance the way information is represented.
- **Pattern Context:** With multiple contexts, chatbots must be able to identify when users switch context. It is important for product designers to intuitively understand what a first-time user may look like, and how each user's learning curve is unique. This design philosophy will optimise the product around how user can, want or need to use a product, rather than forcing them to change their behaviour to accommodate the product. There is no one-size-fits-all approach when it comes to designing the correct conversation flow and currently there are no patterns that work for all chatbot scenarios and domains. Most bots lack the structure to steer a conversation from the beginning to the end. Bots personality has to defines steps to guide user to learn and manage their intention. Moreover, the decision three in bot conversation should lead to results fulfilling these steps. The bot should always be one step ahead of the user. For example, the bot should state which topics it covers when greeting the user.
- **Logical Flow:** Conversation interfaces can provide the opportunity for the user to state what he/she wants to do in his/her own terms, just as he/she would do to another person, and the system takes care of the complexity [4]. The chatbot should have logical thinking that fits with its domain, tasks and behaviour. Users expect the bot to remember information such as their name or topics already covered in conversations. Moreover, they expect more flexibility in terms of interaction-multiple feedback options and more free speech. Finally, they expect the bot to be intelligent and reason about the situation and interaction paradigm. We illustrate a logical conversation flow between the user and a pizza ordering chatbot.

```
Bot: You can select pizza pepperoni, pizza
margarita and pizza with veggies. What is your
choice ?
Human: I want pizza pepperoni
Bot: 1 pizza pepperoni, cool, now tell me the
size (small, medium, large) ?
```

In this scenario the bot asks more details to confirm the users order. The user basically responds to various chatbot questions. Although this example is short, it does show many of the key issues that need to be dealt with in a question-based practical dialogue.

- **Linguistic Constraints:** This refers to the way the chatbot builds a communication with the user. There are several approaches to this issue, one of the mainstay techniques is the semantic restrictions in the grammar to enforce semantic, as well as syntactic, constraints. For example, we might encode a restriction that the verb eat applies to objects that are edible. However, if we want to allow all possible sentences in a conversation, then it becomes hard to find all semantics. The other technique used in this domain is the intend detection or recognition. This is strongly related to NLP, especially with the introduction of statistical techniques trained on large corpora, this techniques provides ways to build dialogue systems. However, we should determining what the user is trying to do by saying the utterance. For example, if the user tell the system "I really like vegetable salad", then the system should interpret it as the user is satisfied with the current meal recommendation and interpret it as:

```
User <Sentiment> <Food>
```

All the above domains determine the intelligence of the chatbot system. One has to consider integrating one or more, if not all of the above foundations when analysing, designing and implementing a conversational user interface. A core part of a conversational UI lies in the actual conversation and the ability of a bot to individually and contextually communicate to a humans. A conversational UI reproduces a one-to-one communication between two parties. It seems relevant to look into the fields of communication theory and psychology, behaviour change, machine learning, linguistics, and even logic as there are several models to map communication, conversations and the way they work and function.

## IV. FUTURE INSIGHTS

### A. Approach Consideration

It is important to build a connection between the theory and technology of the chatbot. Current research is not considering major areas in bot design and development. Some major areas include, Chatbots Personality, Flexibility in Response, Simplicity in Interaction, Tasks and Duty Specification, Empathy & Emotional State, User Boredom, to mention a few. The research in this domain is still growing and getting the right piece of information takes a lot of effort. It is essential to follow a pipeline and determine the necessary elements for designing a chatbot with a specific focus domain. In Figure-1 we provide an overview of the step by step pipeline to follow when designing and developing a chatbot system.

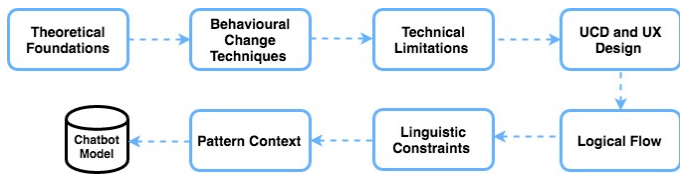

Figure 1. The Pipeline Design for Chatbot Systems.

### B. Scenario Description

Soon, conversational-AI will substitute and leave many technologies obsolete. Depending on the context, they will be to get you that exact right piece of content. To better illustrate what are the necessary steps to follow when designing and developing such system, we provide a scenario describing how developer can follow the pipeline above to develop a chatbot with validated approach. Consider developing a chatbot for health dietary adherence and Mathew, the chatbot developer.

*Mathew is planning to develop a chatbot system that is simple to use, effective and responds well to user requests. He decides to follow the chatbot system pipeline to cover the behavioural, theoretical, technical and other aspects given by the pipeline. Mathew starts by checking the theoretical foundations for the chatbot system design (e.g., theory to motivate vegetable consumption). This step will include understating the theory behind user motivation, effective intervention by the bot and long-term adherences. Now Mathew has to decide and choose among the available theoretical models. After choosing the theoretical model to follow, Mathew has to check for the behavioural change technique to develop into the bot and achieve the initial goal set (e.g., a framework to implement the behaviour of vegetable consumption). Next, Mathew has to start the UX design, it seems trivial, however, this step can get extremely complex specially with chatbot systems. Mathew has to follow a UX design concept that better works for chatbot systems. He has to choose among design functionalities to embed in the chat UI (e.g., whether allow user to type or provide them with buttons to report their daily dietary activities). Now that Mathew has decided the theory and technique to implement the theory and he designed the initial CUI, he has to perform analysis and requirement engineering on the technical tools and models he has to use to power the bot with AI and decide the necessary intelligent behaviour the bot has to carry out. Mathew has to decide whether he will use rule-based or AI-based approach to implement various behaviours into the chatbot. Moreover, he can use some available platforms to add more behaviour oriented functionalities (e.g., perform sentiment analysis and intend detection). Finally, he will have to well craft the pattern that better fits the context of the chatbot and the logical flow of the conversation, taking into account the personality and language of the chatbot.*

## V. STUDY LIMITATIONS

This is a preliminary study on the challenges to account for when designing and developing chatbot systems intended for lifestyle promotion. We did not discuss in details the theoretical models, behavioural change techniques, the available machine learning and neural network systems or APIs that support building conversational user interfaces. Moreover, although we provided an insight about the UX design for CUI, however, we did not provide a comprehensive description of the most important design elements to consider when designing CUIs. Future work should consider developing domain specific models that encompasses behavioural, theoretical, and technical models to serve a specific chatbot system. Also, when working on the NLP and linguistic of the chatbot, a special attention has to be given to integrating domain specific information and avoiding general purpose models. In this stage, developing and validating first basic prototypes can provide impression of chatbot usability and effectiveness.

## VI. Conclusion

The use of chatbots within the field of health promotion has a large potential to reach a varied group of people. This paper described the background and motivation for chatbot system in the context of healthy nutrition recommendation. We discussed current challenges associated with chatbot system, we tackled technical, theoretical, behavioural, and social aspects of the challenges. A pipeline has been proposed which developers can follow as guidelines when designing and implementing such systems. This will ensure the system compliance with the pipeline goals. A lot of work has to be done on the challenges discussed to get deeper insight into the challenges, and hence improve the existing chatbot system design and implementation. Although there are still serious technical challenges to overcome, chatbot systems are showing promises. Next, we plan to develop a prototype system reflecting on all the challenges discussed, and following the pipeline steps. We will validate this with an experiment with users struggling with poor nutrition and dietary challenges. This will provide a more significant assessment of the approach.